\pdfoutput=1
\documentclass[10pt,twocolumn,letterpaper]{article}

\usepackage[pagenumbers]{wacv} % To force page numbers, e.g. for an arXiv version

\usepackage{graphicx}
\usepackage{amsmath}
\usepackage{amssymb}
\usepackage{booktabs}
\usepackage{xcolor}
\usepackage[accsupp]{axessibility}  % Improves PDF readability for those with disabilities.
\usepackage[misc]{ifsym}

\newcommand*{\affaddr}[1]{#1} % No op here. Customize it for different styles.
\newcommand*{\affmark}[1][*]{\textsuperscript{#1}}
\newcommand*{\email}[1]{\texttt{#1}}

\usepackage[pagebackref,breaklinks,colorlinks]{hyperref}

\usepackage[capitalize]{cleveref}
\crefname{section}{Sec.}{Secs.}
\Crefname{section}{Section}{Sections}
\Crefname{table}{Table}{Tables}
\crefname{table}{Tab.}{Tabs.}

\def\wacvPaperID{296} % *** Enter the WACV Paper ID here
\def\confName{WACV}
\def\confYear{2024}

\begin{document}

% \title{Exploration of Text-to-Image Diffusion Models \\ to Customize 360-degree Panoramas}
\title{Customizing 360-Degree Panoramas through Text-to-Image Diffusion Models}
% add more experimental results in the supplementary material, ablation study for sliding distance, more qualitative results.

% \author{Hai~Wang\\
% University College London\\
% {\tt\small hai.wang.22@ucl.ac.uk}
% \and
% Xiaoyu Xiang\\
% Meta Reality Labs\\
% {\tt\small xiaoyu.xiang.ai@gmail.com}
% \and
% Yuchen Fan\\
% Meta Reality Labs\\
% {\tt\small fyc0624@gmail.com}
% \and
% Jing-Hao Xue\\
% University College London\\
% {\tt\small jinghao.xue@ucl.ac.uk}
% }

\author{%
Hai~Wang\affmark[1]\thanks{Corresponding author. All experiments, data collection and processing were conducted in University College London.} \quad Xiaoyu Xiang\affmark[2] \quad Yuchen Fan\affmark[2] \quad Jing-Hao Xue\affmark[1]\\
\affaddr{\affmark[1]University College London} \quad \affaddr{\affmark[2]Meta Reality Labs}\\
\email{\tt\small \{hai.wang.22, jinghao.xue\}@ucl.ac.uk}\tt\small, \email{\tt\small \{xiangxiaoyu, ycfan\}@meta.com}%
}

\maketitle

%%%%%%%%% ABSTRACT
\begin{abstract}
Personalized text-to-image (T2I) synthesis based on diffusion models has attracted significant attention in recent research. However, existing methods primarily concentrate on customizing subjects or styles, neglecting the exploration of global geometry. In this study, we propose an approach that focuses on the customization of 360-degree panoramas, which inherently possess global geometric properties, using a T2I diffusion model. To achieve this, we curate a paired image-text dataset specifically designed for the task and subsequently employ it to fine-tune a pre-trained T2I diffusion model with LoRA. Nevertheless, the fine-tuned model alone does not ensure the continuity between the leftmost and rightmost sides of the synthesized images, a crucial characteristic of 360-degree panoramas. To address this issue, we propose a method called StitchDiffusion. Specifically, we perform pre-denoising operations twice at each time step of the denoising process on the stitch block consisting of the leftmost and rightmost image regions. Furthermore, a global cropping is adopted to synthesize seamless 360-degree panoramas. Experimental results demonstrate the effectiveness of our customized model combined with the proposed StitchDiffusion in generating high-quality 360-degree panoramic images. Moreover, our customized model exhibits exceptional generalization ability in producing scenes unseen in the fine-tuning dataset. Code is available at \href{https://github.com/littlewhitesea/StitchDiffusion}{https://github.com/littlewhitesea/StitchDiffusion}.
\end{abstract}

%%%%%%%%% BODY TEXT
\section{Introduction}
\label{sec:intro}

\begin{figure*}[t]
\centering
\vspace{-7mm}
\includegraphics[height=3.8cm]{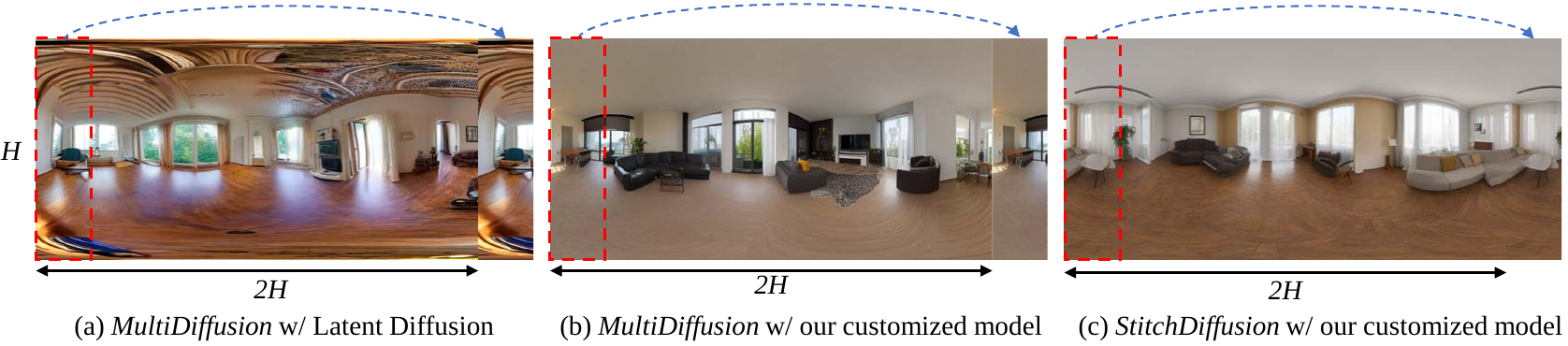}
\caption{Example results of three different methods with the text prompt `\emph{\textcolor{red}{$V^*$}, a living room with a couch and a table}', where $V^*$ refers to the trigger word. To easily recognize the continuity or discontinuity between the leftmost and rightmost sides of the generated image, we copy the leftmost area indicated by the \textcolor{red}{red dashed box} and paste it onto the rightmost side of the image. The notation `w/' denotes `with'. Our customized model is achieved by fine-tuning latent diffusion . Compared with \emph{MultiDiffusion} \cite{bar2023multidiffusion} combined with latent diffusion  in (a), \emph{MultiDiffusion} \cite{bar2023multidiffusion} with our customized model in (b) generates more visually appealing content. Moreover, in contrast to (b), our proposed \emph{StitchDiffusion} with the customized model in (c) successfully synthesizes a seamless 360-degree panoramic image.}
\vspace{-5mm}
\label{fig:teaser}
\end{figure*}

360-degree panoramic images \cite{pano3,pano2,pano4,pano1} are extensively utilized in virtual reality (VR) devices, such as head mount displays \cite{anthes2016state}. Unlike ordinary two-dimensional (2D) images, which have a limited viewing range, 360-degree panoramas encompass the entire $360^{\circ}\times180^{\circ}$ field of view. This allows viewers to explore a scene from any angles, providing them with an immersive experience. The inherent globally geometric properties of 360-degree panoramas stem from this unique characteristic. There are various types of projections \cite{xu2020state} used to represent 360-degree panoramas. In this paper, we specifically focus on the equirectangular projection (ERP), which represents the 360-degree panoramic image on a 2D surface. In this context, two essential properties of a 360-degree panorama arise: (1) the width of a 360-degree panoramic image is twice its height, and (2) the leftmost and rightmost sides of a 360-degree panorama are continuous.

Diffusion models \cite{croitoru2023diffusion,dhariwal2021diffusion,yang2022diffusion} perform better in generating photorealistic and diverse images compared with generative adversarial networks (GANs) \cite{creswell2018generative,goodfellow2020generative}, leading to increasing attention over the past two years. Thanks to their excellent generation quality and controllability, diffusion models have been widely explored for tackling numerous challenging tasks \cite{avrahami2022blended,graikosdiffusion,ho2022imagen,li2022diffusion,nie2022diffusion,latentdiffusion,vahdat2022lion}. Notably, diffusion models applied to text-to-image (T2I) synthesis \cite{glide,dall-e2,latentdiffusion,imagen} can produce high-quality images corresponding to descriptive text prompts, making them highly popular on social media. However, these models have limitations when it comes to synthesizing instances of customized concepts, such as a user's pet or personal item. 

To handle this challenge, several personalized T2I generation algorithms \cite{textualinversion,svdiff,customdiffusion,dreambooth,key-locked} have been proposed. These algorithms enable the customization of T2I diffusion models by providing multiple images of a specific subject or concept, resulting in the synthesis of images containing the subject or concept in diverse contexts. Different from these existing personalized technologies \cite{textualinversion,customdiffusion,dreambooth} which focus on customizing specific subjects (e.g., dog, sunglasses) or styles (e.g., oil painting, pop art), our work aims to explore the customization of global geometry.

Specifically, we focus on customization of a T2I diffusion model for synthesizing 360-degree panoramas with inherent globally geometric properties. To begin, we build a paired image-text dataset called \emph{360PanoI}. Due to limited computational resources and the need for fine-tuning efficiency, we employ the Low-Rank Adaptation (LoRA) \cite{lora,lorat2i} technology to fine-tune a pre-trained T2I diffusion model using the collected \emph{360PanoI} dataset. However, we encounter difficulties when generating 360-degree panoramas using the fine-tuned diffusion model, even with the use of \emph{MultiDiffusion} \cite{bar2023multidiffusion}, a recent method for producing traditional panoramas but disregarding the continuity between the leftmost and rightmost sides of the synthesized image (as shown in Figure \ref{fig:teaser}). To address this issue, we put forward a tailored generation process named \emph{StitchDiffusion} for synthesizing 360-degree panoramas. In the \emph{StitchDiffusion} approach, we perform pre-denoising operations twice at each time step of the denoising process on the stitch block, which is constituted of the leftmost and rightmost image regions. After the denoising process is completed, we conduct a global cropping to produce the final image. This method guarantees that the fine-tuned T2I diffusion model generates seamless 360-degree panoramic images. Moreover, despite the limited number of scenes in our \emph{360PanoI} dataset, the fine-tuned diffusion model demonstrates excellent generalization capabilities to unseen scenes. In other words, the fine-tuned diffusion model can successfully synthesize 360-degree panoramas of scenes not present in the fine-tuning dataset. This observation indicates that T2I diffusion models possess the potential to effectively capture and represent global geometry.

The contributions of this work can be summarized as follows: (1) We make the first attempt to explore the customization of 360-degree panoramas using T2I diffusion models, which is beneficial for employing T2I diffusion models in various application scenarios, such as indoor design and VR content creation. Our experimental results demonstrate that T2I diffusion models possess the capability to produce 360-degree panoramas with inherent geometric properties and generalize this ability to unseen scenes. (2) We propose a stitch method called \emph{StitchDiffusion} as part of the generation process to synthesize seamless 360-degree panoramic images, which ensures the continuity between the leftmost and rightmost sides of the synthesized panoramas. (3) We curate a paired image-text dataset called \emph{360PanoI} specifically for the synthesis of 360-degree panoramas. This dataset serves as a valuable resource for future studies and advancements in the field of 360-degree panoramic images.

%------------------------------------------------------------------------

\section{Related Work}

\noindent
\textbf{Text-to-Image Diffusion Models.}
Text-to-image (T2I) synthesis based on diffusion models \cite{gu2022vector,glide,dall-e2,latentdiffusion,imagen,latentdiffusion} can generate images that align with the provided text prompts, which have showcased unprecedented levels of diversity and fidelity. We will only introduce several representative works here; for more comprehensive information, we refer readers to the survey paper \cite{t2isurvey}. GLIDE \cite{glide} stands out as a pioneering T2I diffusion model that uses classifier-free guidance in the T2I synthesis process. Different from GLIDE requiring to train its text encoder, Imagen \cite{imagen} utilizes a pre-trained large transformer language model to encode textual input for image generation. Both GLIDE and Imagen operate in the pixel space, which demands substantial computational resources. To alleviate this requirement, latent diffusion Models (LDMs)  propose to train diffusion model in the latent space, significantly reducing the computational burden. In our work, we adopt LDM  as the foundational model due to its relatively lower demand for computing resources.

\noindent
\textbf{Personalized Text-to-Image Generation.}
Given one or multiple images of a specific subject or style provided by users, personalized text-to-image (T2I) generation \cite{neti,textualinversion,e4t,svdiff,customdiffusion, dreambooth,instantbooth,c-lora,key-locked,eti,elite} based on diffusion models aims to synthesize instances of  the specific subject or style in diverse contexts. These personalized techniques can be broadly categorized into three groups. The first category is  the \emph{personalization-by-inversion} approach, initially explored in Textual Inversion \cite{textualinversion}. This method optimizes an input vector in the textual embedding space to represent the desired subject or style. To enhance its expressive power, Extended Textual Inversion \cite{eti} and NeTI \cite{neti} propose optimizing multiple vectors and employing a neural mapper, respectively, resulting in stronger representations. DreamBooth \cite{dreambooth}, on the other hand, is a pioneering \emph{personalization-by-fine-tuning} method. It introduces a class-specific prior preservation loss to mitigate language drift \cite{languagedrift1,languagedrift2}, and fine-tunes the entire T2I diffusion model for binding unique identifiers to user-provided subjects. In contrast, Custom Diffusion \cite{customdiffusion} and SVDiff \cite{svdiff} fine-tune only a small portion of parameters for improved efficiency. However, these approaches still require fine-tuning the diffusion model for each user-specific subject. Recognizing this limitation, \emph{personalization-by-encoder} methods \cite{e4t,instantbooth,elite} have been proposed for rapid customization of T2I models. Specifically, these methods first train a mapping encoder, which is then used to directly map arbitrary input images into word embeddings representing the subject. Unlike the existing personalized approaches that concentrate on customizing specific subjects or artistic styles, our work in this paper explores the customization of global geometry, specifically 360-degree panoramic images. The successful customization of such complex geometries would demonstrate the inherent ability of T2I diffusion models to capture intricate spatial representations.

\noindent
\textbf{Panorama Generation.} GAN-based panorama generation algorithms \cite{inout,coco,infinitygan,bips,boundless,stylelight,wu2022cross} have been extensively studied. In contrast, Text2Light \cite{text2light} adopts a text-conditioned global sampler and structure-aware local sampler to generate panoramic images by sampling from a dual-codebook representation. Recently, diffusion models have also shown promising results in panorama synthesis \cite{bar2023multidiffusion,panogen,diffcollage}. DiffCollage \cite{diffcollage} utilizes a semantic segmentation map as the condition for the diffusion model and generates 360-degree panoramas based on a complex factor graph. On the other hand, PanoGen \cite{panogen} employs latent diffusion  to synthesize new indoor panoramic images with a recursive image outpainting technology based on multiple text descriptions. Distinguishing itself from PanoGen \cite{panogen}, \emph{MultiDiffusion} \cite{bar2023multidiffusion} simultaneously samples the panoramic image through blending diffusion paths of all overlapped cropped patches to synthesize high-quality images. However, \emph{MultiDiffusion} \cite{bar2023multidiffusion} does not guarantee the continuity between the leftmost and rightmost sides of the generated image, which is a natural property of the 360-degree panorama. To deal with this problem, we propose in this paper a method called \emph{StitchDiffusion}. This approach leverages our customized diffusion model to synthesize panoramas that exhibit continuity between the leftmost and rightmost sides, resulting in a seamless viewing experience. Moreover, we demonstrate in this paper that our customized diffusion model possesses strong generalization capabilities, allowing it to generate a wide range of 360-degree panoramas in various contexts, even for scenes not present in the fine-tuning dataset.
%-------------------------------------------------------------------------

\begin{figure*}[t]
\centering
\vspace{-5mm}
\includegraphics[height=8.5cm]{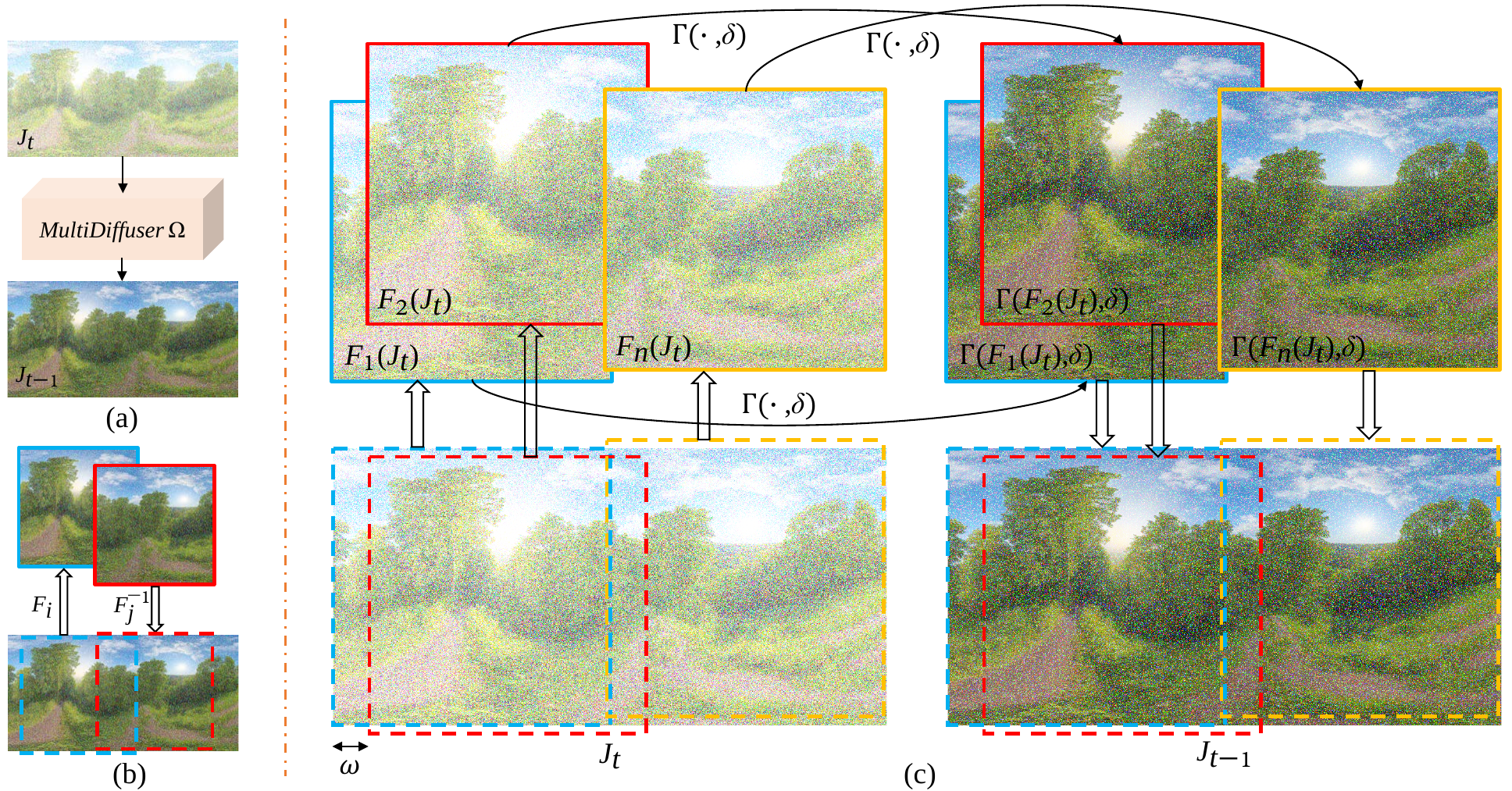}
\caption{Overview of \emph{MultiDiffusion} \cite{bar2023multidiffusion} for panorama generation. (a) The input $J_t$ and output $J_{t-1}$ of the \emph{MultiDiffuser} $\Omega$ at time step $t$ during the denoising process. (b) Illustration of the mapping $F_i$ (directly cropping a patch from an image) and inverse mapping $F^{-1}_j$, where $i,j\in\left\{1, 2, \cdots, n\right\}$. (c) Detailed inner process of the \emph{MultiDiffuser} $\Omega$ at time step $t$ during the denoising process. Here, $\Gamma(\cdot,\delta)$ denotes the pre-trained T2I diffusion model with the textual embedding $\delta$ from a given text prompt, and $\omega$ is the horizontal sliding distance between adjacent cropped patches. Note that \emph{MultiDiffusion} cannot guarantee the continuity between the leftmost and rightmost sides of generated panoramic images.}
\vspace{-3mm}
\label{fig:multi}
\end{figure*}

\begin{figure}[t]
\centering
\vspace{-3mm}
\includegraphics[height=1.85cm]{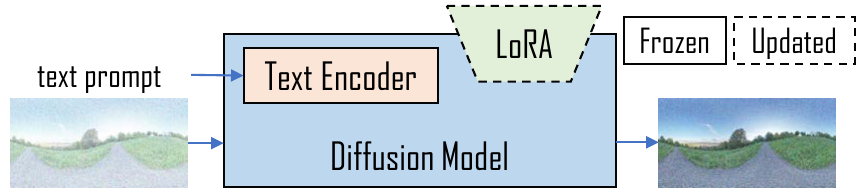}
\caption{Illustration of customizing a T2I diffusion model with LoRA \cite{lora} for synthesizing 360-degree panoramic images. The paired image-text data used during the fine-tuning process are from our collected \emph{360PanoI} dataset.}
\vspace{-3mm}
\label{fig:lora}
\end{figure}

\begin{figure*}[t]
\centering
\vspace{-5mm}
\includegraphics[height=7.3cm]{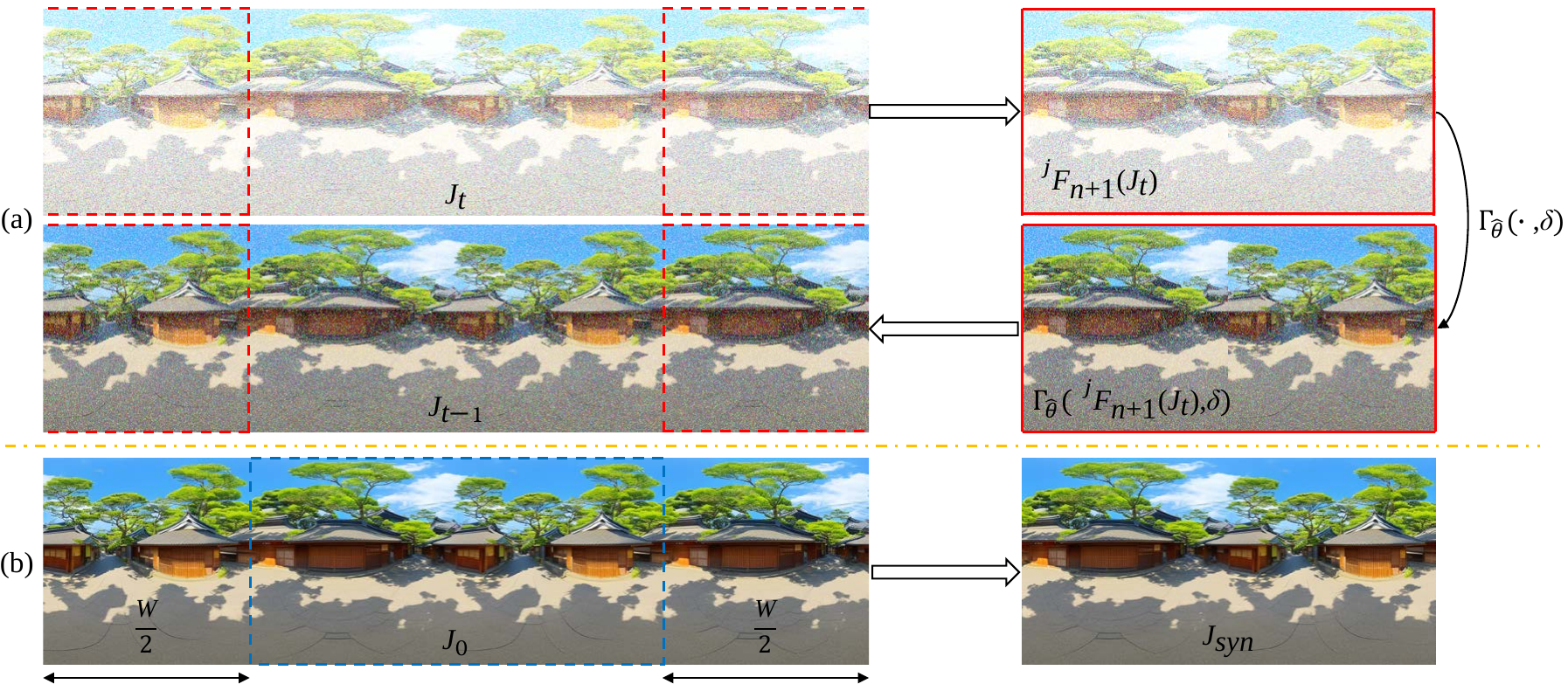}
\caption{Overview of our proposed \emph{StitchDiffusion} for generating 360-degree panoramas. (a) At each time step $t$ of the denoising process, the $H\times{W}$ stitch block undergoes pre-denoising operations twice, which is constituted by the leftmost ($H\times{\frac{W}{2}}$) and rightmost ($H\times{\frac{W}{2}}$) regions of the image $J_t$. Here, the value of $W$ is twice that of $H$. (b) The global cropping denoted by the \textcolor{blue}{blue dashed box} of the final clear result $J_0$ is taken to achieve the 360-degree panorama $J_{syn}$. Note that if image $J_0$ ($H\times4H$) is divided horizontally along the middle into two equal parts, then the left half ($H\times2H$) of $J_0$ is identical to the right half ($H\times2H$) of $J_0$, which could ensure the continuity between the leftmost and rightmost sides of the $J_{syn}$ obtained from global cropping. }
\vspace{-5mm}
\label{fig:stitch}
\end{figure*}

\section{Methodology}

In this section, we first briefly review the use of \emph{MultiDiffusion} \cite{bar2023multidiffusion} to synthesize panoramic images. Then, we describe the process of customizing a pre-trained T2I diffusion model for 360-degree panoramas. Finally, we introduce our proposed method, \emph{StitchDiffusion}, which is able to handle the issue of discontinuity between the leftmost and rightmost sides of the generated 360-degree panoramic image when using \emph{MultiDiffusion} \cite{bar2023multidiffusion}. 

\subsection{Preliminaries}

Given a pre-trained T2I diffusion model $\Gamma$, the sequential denoising process of this model, gradually from a noisy image $I_T$ to the final clean image $I_0$, could be expressed as
\begin{equation}\label{eq:ref_model}
I_{t-1} = \Gamma(I_{t},\delta)\ , t\in\left\{T,T-1,\cdots,1\right\} \ ,
\end{equation}
where $I$ is in the image space $\mathcal{I}=\mathbb{R}^{H\times{W}\times{C}}$, and $\delta$ is the textual embedding of a text prompt. A target of \emph{MultiDiffusion} \cite{bar2023multidiffusion} is to generate a panoramic image aligning with the given text prompt without the need for any training or fine-tuning of the diffusion model $\Gamma$, which serves as a reference model. To this end, \emph{MultiDiffusion} \cite{bar2023multidiffusion} defines a different T2I diffusion model $\Omega$ called \emph{MultiDiffuser}, which operates in an image space $\mathcal{J}$. Its sequential denoising process is
\begin{equation}\label{eq:multi_model}
J_{t-1} = \Omega(J_{t},\delta)\ , t\in\left\{T,T-1,\cdots,1\right\} \ ,
\end{equation}
where $J$ is in the image space $\mathcal{J}=\mathbb{R}^{H\times{W^{\prime}}\times{C}}$, and $W^{\prime}$ is greater than or equal to $W$. 

To establish a connection between the target image space $\mathcal{J}$ and reference image space $\mathcal{I}$, \emph{MultiDiffusion} \cite{bar2023multidiffusion} further defines $n$ mappings $F_i: \mathcal{J}\rightarrow{\mathcal{I}}$, where $i\in\left\{1, 2, \cdots, n\right\}$. At time step $t$ of the denoising process, $F_i(J_t)\in{\mathcal{I}}$ is the $i$-th $H\times{W}$ cropped patch of image $J_t$. These $n$ overlapped cropped patches cover the whole image $J_t$, illustrated in Figure \ref{fig:multi}\textcolor{red}{(c)}. The value of $n$ is determined by
\begin{equation}\label{eq:crop_num}
n = \frac{W^{\prime}-W}{\omega} + 1 \ ,
\end{equation}
where $\omega$ denotes the horizontal sliding distance between adjacent cropped patches.
Using these mappings, the new denoising process of diffusion model $\Omega$ at time step $t$ is achieved by solving the following optimization problem:
\begin{equation}\label{eq:optimization}
\Omega(J_t,\delta) = \mathop{\arg\min}\limits_{J\in\mathcal{J}} \sum_{i=1}^{n}\Vert F_i(J) - \Gamma(F_i(J_t),\delta) \Vert^2 \ .
\end{equation}
In fact, this is a least-square problem and the corresponding closed-form solution is 
\begin{equation}\label{eq:closed-form}
\Omega(J_t,\delta) = \sum_{i=1}^{n}\frac{F^{-1}_i(\mathbf{1})}{\sum_{j=1}^{n}F_{j}^{-1}(\mathbf{1})} \otimes F_i^{-1}(\Gamma(F_i(J_t),\delta)) \ ,
\end{equation}
where $\mathbf{1}$ is in the image space $\mathcal{I}=\mathbb{R}^{H\times{W}\times{C}}$, and its all pixel values are equal to 1. By leveraging this new denoising process, \emph{MultiDiffuser} $\Omega$ can directly utilize the pre-trained T2I diffusion model $\Gamma$ without any training or fine-tuning steps to generate panoramic images aligned with the given text prompt.

\subsection{Customizing Models with LoRA}

To customize a pre-trained T2I diffusion model for 360-degree panorama synthesis, we start by collecting a dataset of 360-degree panoramic images. Then, these images are tagged using BLIP \cite{blip}, creating a paired image-text dataset called \emph{360PanoI}. More detailed information about the collected dataset can be found in Section \ref{360pano}. For the fine-tuning process, we employ Low-Rank Adaptation (LoRA) \cite{lora} technology, which was initially proposed for fine-tuning large language models. Specifically, LoRA introduces trainable rank decomposition matrices into the pre-trained model, allowing for faster adaptation to downstream tasks with lower computational requirements compared to full fine-tuning. Recent work \cite{lorat2i} has validated the effectiveness of LoRA in pre-trained T2I diffusion models. Considering its efficiency and low demand for computational resources, we employ LoRA to fine-tune the pre-trained diffusion model for generating 360-degree panoramic images using the \emph{360PanoI} dataset, as shown in Figure \ref{fig:lora}.

Given the ground-truth image $I^{gt}$ and its corresponding textual embedding $\delta$, the preliminary customized model $\Gamma_{\theta}$ with LoRA is fine-tuned by using the loss function $L_{pano}$ to denoise a variably-noised image $\alpha_t I^{gt} + \sigma_t \boldsymbol{\epsilon}$ as follows:
\begin{equation}\label{eq:loss}
L_{pano} = \mathbb{E}_{I^{gt},\delta,\boldsymbol{\epsilon},t} \left[\gamma_t\Vert \Gamma_{\theta}(\alpha_t I^{gt} + \sigma_t \boldsymbol{\epsilon}, \delta) - I^{gt} \Vert^2\right] \ ,
\end{equation}
where $\theta$ refers to the trainable matrices of LoRA, $\boldsymbol{\epsilon}$ and $\gamma_t$ represent the noise following a Gaussian distribution and the functions of diffusion process time $t$, respectively, and $\alpha_t$ and $\sigma_t$ are terms used to manage the noise schedule and the sample quality, respectively. Upon completing the fine-tuning process, we obtain the final customized diffusion model denoted as $\Gamma_{\hat{\theta}}$, where $\hat{\theta}$ denotes the updated parameters of LoRA.

\subsection{StitchDiffusion for 360-degree Panoramas}

Firstly, let us review two natural properties of a 360-degree panorama represented by the equirectangular projection \cite{xu2020state}: (1) the width $W$ of the 360-degree panoramic image is twice its height $H$, resulting in a final generated panorama size of $H\times{2H}$; (2) there should be continuity between the leftmost and rightmost sides of the 360-degree panorama. However, as shown in Figure \ref{fig:multi}, the \emph{MultiDiffusion} method fails to ensure this continuity. To solve this problem and guarantee seamless 360-degree panoramas, we put forward a new generation process called \emph{StitchDiffusion}.

Specifically, we utilize the \emph{MultiDiffuser} $\Omega$ to generate a panoramic image $J$ with a resolution of $H\times{(2H+2H)}$ using the customized diffusion model $\Gamma_{\hat{\theta}}$, that is, $W^{\prime}$ in Equation \ref{eq:crop_num} is equal to $4H$. In addition, for \emph{MultiDiffuser} $\Omega$ at time step $t$ of denoising process in Figure \ref{fig:multi}, we additionally employ the customized diffusion model $\Gamma_{\hat{\theta}}$ to perform pre-denoising operations twice on a stitch block, which consists of the leftmost ($H\times\frac{W}{2}$) and rightmost ($H\times\frac{W}{2}$) regions of the current noisy image $J_t$, as illustrated in Figure \ref{fig:stitch}\textcolor{red}{(a)}. Here, $W$ is twice the value of $H$. In this situation, the corresponding denoising process at time step $t$ of our proposed \emph{StitchDiffusion} can be expressed as
\begin{equation}\label{eq:stitch}
\begin{split}
J_{t-1} = \sum_{j=1}^{2}\frac{{^{j}F^{-1}_{n+1}}(\mathbf{1})}{\Lambda} \otimes {^{j}F^{-1}_{n+1}}(\Gamma_{\hat{\theta}}({^{j}F_{n+1}}(J_t),\delta)) \\ + \sum_{i=1}^{n}\frac{F^{-1}_i(\mathbf{1})}{\Lambda} \otimes F_i^{-1}(\Gamma_{\hat{\theta}}(F_i(J_t),\delta))
  \ ,
\end{split}
\end{equation}
where ${^{j}F_{n+1}}(\cdot)$ and ${^{j}F}^{-1}_{n+1}(\cdot)$ denote the $j$-th additional mapping and inverse mapping of the stitch block, respectively, $\Lambda$ denotes ${{^{1}{F}}_{n+1}^{-1}(\mathbf{1}) + {^{2}{F}}_{n+1}^{-1}(\mathbf{1}) + \sum_{j=1}^{n}F_{j}^{-1}(\mathbf{1})}$. Using the denoising process of our \emph{StitchDiffusion}, we can get a clear image $J_0$ with a resolution of $H\times{(2H+W)}$ at the end of the entire denoising process. To obtain the final 360-degree panoramic image $J_{syn}$  with a resolution of $H\times{2H}$, we perform a global cropping operation on $J_0$:
 \begin{equation}\label{eq:global_crop}
J_{syn} = J_0{[\frac{W}{2}:-\frac{W}{2}]} \ ,
\end{equation}
illustrated in Figure \ref{fig:stitch}\textcolor{red}{(b)}. This operation ensures that the leftmost and rightmost sides of the panoramic image $J_{syn}$ are continuous, as desired for a 360-degree panorama.

%-------------------------------------------------------------------------

\begin{figure*}[t]
\centering
\vspace{-7mm}
\includegraphics[height=9.9cm]{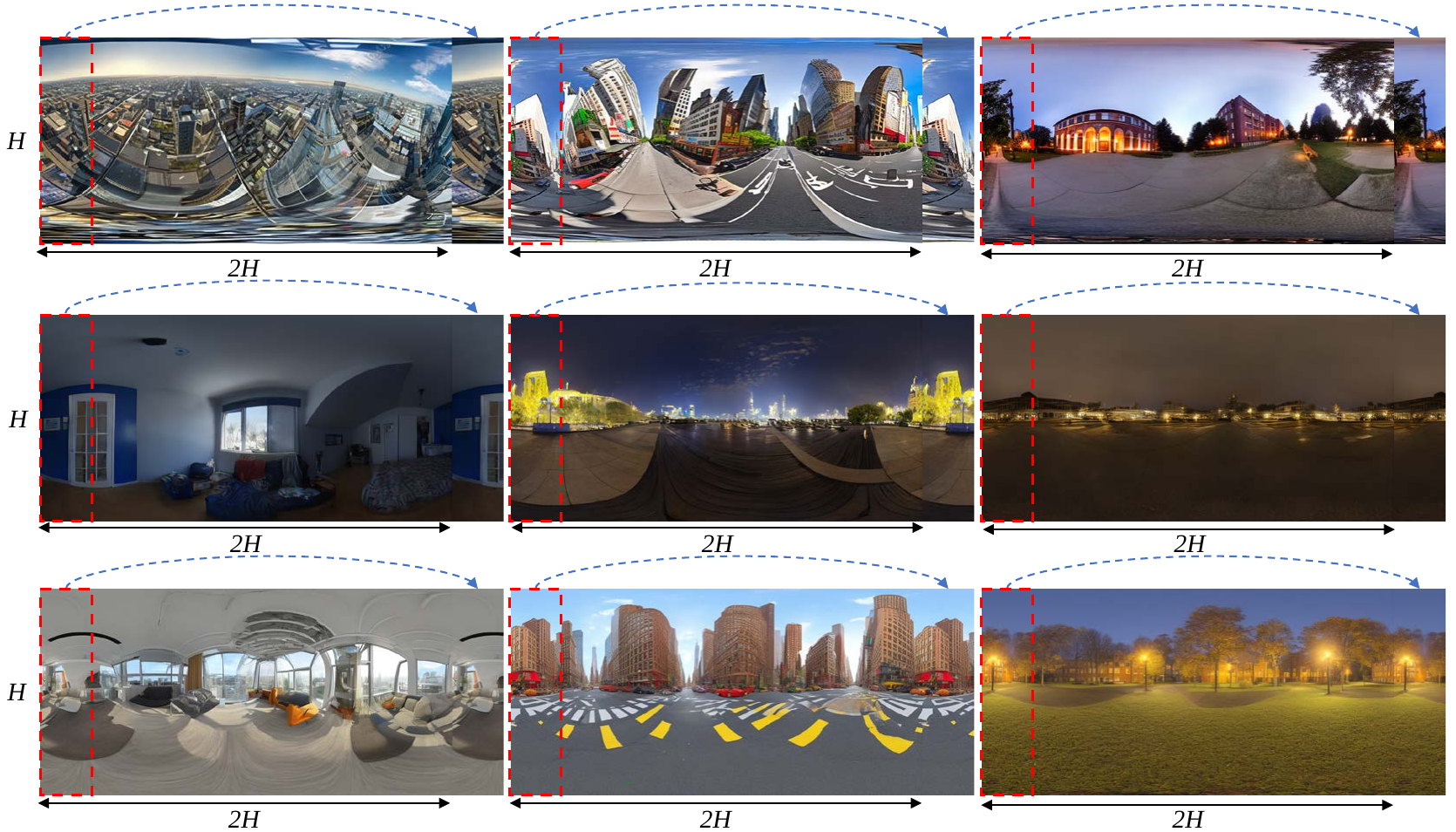}
\caption{Visual comparison among \emph{MultiDiffusion} \cite{bar2023multidiffusion} combined with latent diffusion  (the first row), Text2Light \cite{text2light} (the second row) and our method (the third row). The corresponding text prompts of these images from left to right are `$V^*$, futuristic flat, concept art', `$V^*$, cartoon new york street', and `$V^*$, university campus, foggy night', respectively. To demonstrate the discontinuity or continuity between the leftmost and rightmost sides of the generated image, we copy the left area denoted by the \textcolor{red}{red dashed box} and paste it onto the rightmost side of the image. We can see that our method generates seamless and plausible 360-degree panoramas corresponding to the input text prompts. For more comparison results, please refer to the supplementary material.}
\vspace{-3mm}
\label{fig:comparison}
\end{figure*}

\section{Experiments}

\subsection{Dataset and Implementation Details}

\noindent
\textbf{Dataset.}\label{360pano}
We collected 120 360-degree panoramic images in the real world from Poly Haven \cite{polyhaven}. The images were sourced from a variety of scenes, including \emph{indoor}, \emph{nature}, \emph{night}, \emph{outdoor}, \emph{skies}, \emph{studio}, \emph{sunrise-sunset}, and \emph{urban} settings. Each scene consists of 15 panoramas. Due to limited computational resources, we performed an 8x rescale operation on these images using bilinear interpolation to obtain 360-degree panoramas with a resolution of $512\times1024$ pixels. Subsequently, we utilized BLIP \cite{blip} to tag these processed images. However, the generated text prompts contained poor tags such as `3 6 0 picture', which might potentially impact the fine-tuning process. Therefore, we removed these tags. Additionally, we introduced a trigger word `360-degree panoramic image', denoted as $V^*$ in this paper, into each text prompt. Finally, our \emph{360PanoI} dataset is constituted of 120 360-degree panoramas with a resolution of $512\times1024$ pixels, along with their corresponding text prompts. We present one sample image for each scene in the supplementary material.

\begin{table}[t]
\centering
\vspace{-3mm}
\caption{Quantitative comparison between our method and latent diffusion (LD)  combined with \emph{MultiDiffusion} \cite{bar2023multidiffusion}. Our method consisting of the customized model and \emph{StitchDiffusion} is superior to the baseline in terms of both CLIP-score and FID.}
\begin{tabular}{lcc}
\hline
Method                          & CLIP-score$\uparrow$ & FID$\downarrow$ \\ \hline
LD+\emph{MultiDiffusion} & 0.752$\pm$0.023          &   177.886$\pm$6.478 \\ 
Ours                            &  0.768$\pm$0.005         &   160.960$\pm$6.431  \\ \hline
\end{tabular}
\label{tab:comparison}
\vspace{-5mm}
\end{table}

\noindent
\textbf{Implementation Details.}
To customize a T2I diffusion model for 360-degree panorama synthesis using the \emph{360PanoI} dataset, we employed latent diffusion  and LoRA \cite{lora}. In detail, the LoRA architecture consists of two linear layers with an intermediate dimension of 32. During fine-tuning, we used a batch size of 2 and set the learning rate for the T2I diffusion model to 1e-4. The fine-tuning process was performed for 10 epochs using AdamW \cite{adamw}, which took approximately 40 minutes to complete. In the inference stage, we set the values of $H$ and $W$ to 512 and 1024, respectively, while the horizontal sliding distance $\omega$ between adjacent cropped patches was set to 128 in image space. It is important to highlight that the practical implementation of our \emph{StitchDiffusion} process operates on $J$ and $I$ within latent space. That means the values of $H$, $W$ and $\omega$ in latent space are 64, 128 and 16, respectively. All experiments were conducted on a single Tesla T4 GPU.

\begin{figure*}[t]
\centering
\vspace{-7mm}
\includegraphics[height=4cm]{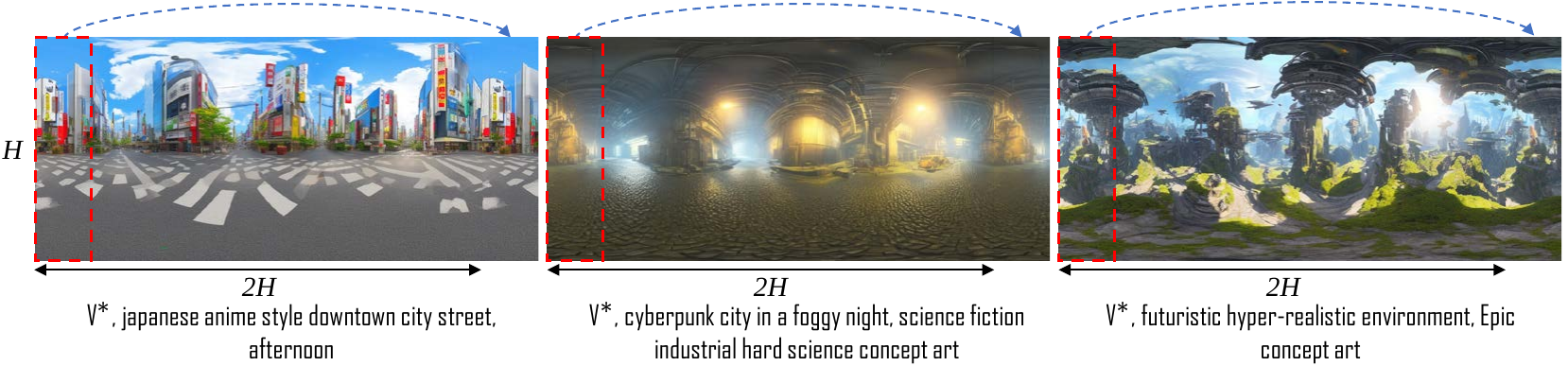}
\caption{Visual results of unseen scenes synthesized by our method. Despite the fact that the collected \emph{360PanoI} dataset only contains 8 scenes from the real world, our customized model with the proposed \emph{StitchDiffusion} effectively produces 360-degree panoramas of diverse unseen scenes, showcasing its excellent generalization ability. More generated images can be seen in the supplementary material.}
\label{fig:generalization}
\end{figure*}

\begin{figure*}[t]
\centering
\vspace{-3mm}
\includegraphics[height=3.7cm]{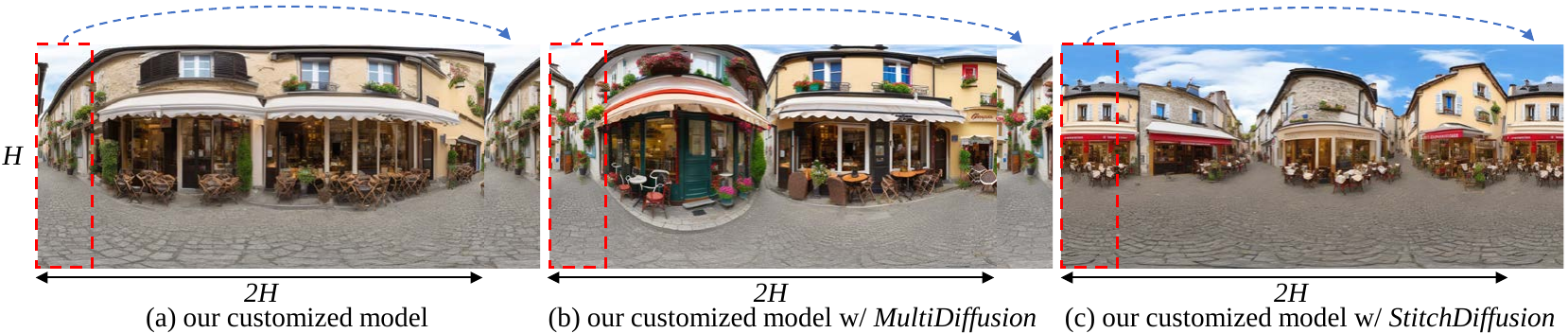}
\caption{Ablation study on \emph{StitchDiffusion}. The corresponding text prompt of these images is `$V^*$, traditional french cafe in the street, small village'. The customized model cannot ensure continuity between the leftmost and rightmost sides of the generated image, even with \emph{MultiDiffusion} \cite{bar2023multidiffusion}. In contrast, \emph{StitchDiffusion} enables the customized model to generate a 360-degree panorama.}
\vspace{-4mm}
\label{fig:ab_stitch}
\end{figure*}

\begin{figure}[t]
\centering
\includegraphics[height=4.9cm]{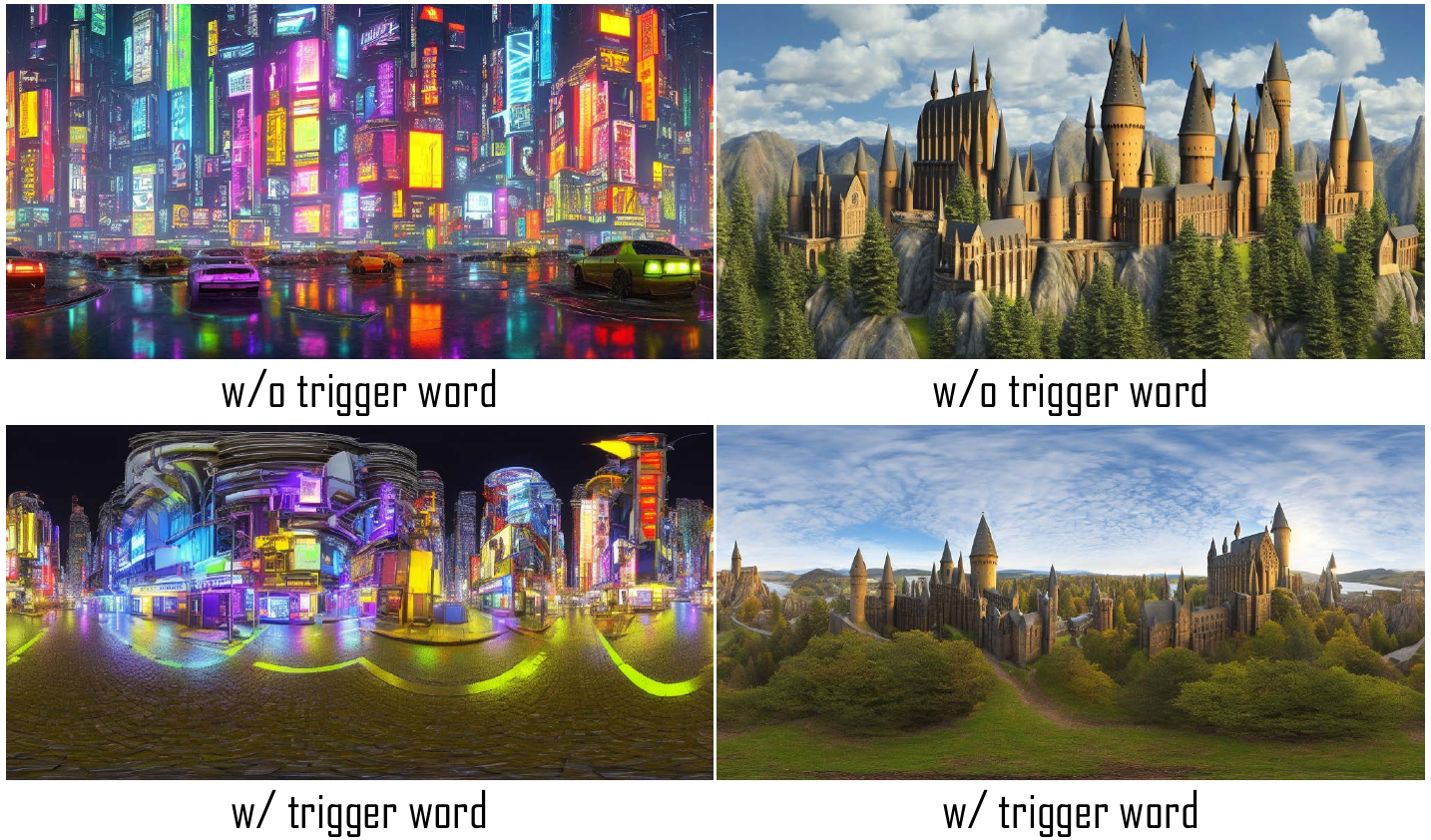}
\caption{Ablation study on the trigger word $V^*$. The corresponding text prompts of left and right images in the second row are `$V^*$, cyberpunk city, neon lights, science fiction' and `$V^*$, magical campus, hyper realistic', respectively. With the trigger word included in the text prompt, the generated images cover the entire $360^{\circ}\times180^{\circ}$ field of view. Here, `w/o' and `w/' refer to `without' and `with', respectively.}
\label{fig:trigger}
\vspace{-5mm}
\end{figure}

\subsection{Comparisons}

Due to the absence of a direct approach based on diffusion model for producing 360-degree panoramic images from an input text prompt, we adopt \emph{MultiDiffusion} \cite{bar2023multidiffusion}, a state-of-the-art method to generate normal panoramas, in combination with latent diffusion  as a baseline. Furthermore, we compare our approach with Text2Light \cite{text2light}, a state-of-the-art non-diffusion-based technique. The visual results are presented in Figure \ref{fig:comparison}. We can see that the baseline method yields images with unsatisfactory content. Notably, the leftmost and rightmost sides of the images generated by the baseline are not continuous, indicating its inability to synthesize 360-degree panoramas. While Text2Light achieves improved continuity in synthesized images, it fails to capture the essence of `futuristic' and `cartoon' themes in the text prompts. In contrast, our proposed method consisting of the customized diffusion model and \emph{StitchDiffusion} produces seamless and plausible 360-degree panoramic images corresponding to the text prompts. 

% The visual results are presented in Figure \ref{fig:comparison}. We can see that the baseline method produces images with unsatisfactory content. In addition, the leftmost and rightmost sides of the images generated by the baseline are not continuous, indicating its inability to synthesize 360-degree panoramas. In contrast, our proposed method consisting of the customized diffusion model and \emph{StitchDiffusion} generates high-quality and seamless 360-degree panoramic images.

To further quantitatively assess the plausibility of images generated by different methods, we first collected additional 20 real 360-degree panoramas from Poly Haven \cite{polyhaven} as our ground truth, and then applied the same processing methodology outlined in Section \ref{360pano} to acquire their text prompts. With these text prompts in hand, we attempted to generate the corresponding images using Text2Light. However, we found that some text prompts exceeded the token limit of Text2Light, preventing Text2Light from synthesizing images with them as inputs. In contrast, both our method and latent diffusion  combined with \emph{MultiDiffusion} \cite{bar2023multidiffusion} can handle all text prompts of the ground truth panoramas. For a fair comparison, we only synthesized the corresponding images using our method and latent diffusion  combined with \emph{MultiDiffusion} \cite{bar2023multidiffusion}.

\begin{figure*}[t]
\centering
\vspace{-5mm}
\includegraphics[height=3.1cm]{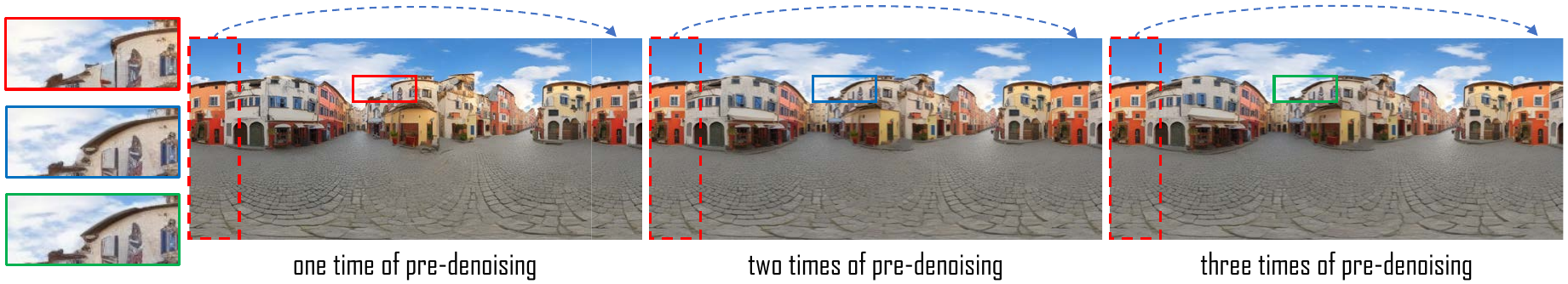}
\caption{Ablation study on the number of pre-denoising times. The corresponding text prompt is `$V^*$, an old city close up, sharp focus'. Although leftmost and rightmost sides of the generated image are continuous when conducting the pre-denoising operation on the stitch block only once at time step $t$ of our \emph{StitchDiffusion}, local content inconsistency in the \textcolor{red}{red solid box} appears in the image. Increasing the number of pre-denoising operations can improve the local content consistency (see the \textcolor{blue}{blue} and \textcolor{green}{green} solid boxes). Considering a trade-off between computational efficiency and image quality, we choose to perform the pre-denoising operations twice on the stitch block.}
\vspace{-5mm}
\label{fig:denoising_time}
\end{figure*}

Now, with these generated images and their corresponding ground truth, we can calculate the quantitative results for the two methods. Specifically, we randomly cropped 1000 patches of size $512\times512$ from the 20 ground truth images and recorded the locations of each patch. Using these recorded locations, we similarly cropped corresponding 1000 patches from the generated images. Next, we employed the image encoder of CLIP \cite{clip} to extract embeddings of these patches. We then calculated the average cosine similarity between the embeddings of the generated patches and the real patches, denoted as CLIP-score. Additionally, we utilized the Frechet Inception Distance (FID) \cite{fid} to quantify the distance between the distribution of generated patches and the distribution of real patches. To further verify the effectiveness and robustness of our method in generating plausible images, we repeated the generation process 10 times, and then calculated the corresponding mean and standard deviation of the two metrics. The quantitative results, as shown in Table \ref{tab:comparison}, indicate that our method outperforms the baseline in terms of the two metrics.

\subsection{Generalizability}

To evaluate the generalization ability of our customized diffusion model to unseen scenes, we fed a variety of text prompts describing scenes not included in the \emph{360PanoI} dataset into the model. The corresponding generated images are presented in Figure \ref{fig:generalization}. It is evident that our customized diffusion model using the proposed \emph{StitchDiffusion} produces visually appealing 360-degree panoramas of diverse unseen scenes, such as Japanese anime style, cyberpunk, and hyper-realistic environment. Remarkably, these results are achieved even though the collected \emph{360PanoI} dataset only contains 8 scenes from the real world, which highlights the excellent generalizability of our customized T2I diffusion model.

\subsection{Ablation Studies}

We only present a subset of our ablation studies here; other ablation studies are in the supplementary material. 

\noindent
\textbf{\emph{StitchDiffusion} Ablation.} To study the effect of our proposed \emph{StitchDiffusion} method on the generated results, we conducted a comparative analysis. Specifically, we compared the images synthesized by our customized diffusion model with and without \emph{StitchDiffusion}. In addition, we introduced the results produced by combining the customized diffusion model with \emph{MultiDiffusion} \cite{bar2023multidiffusion} for a more comprehensive comparison. The generated images are displayed in Figure \ref{fig:ab_stitch}. We can observe that the customized diffusion model alone is unable to synthesize 360-degree panoramas, primarily due to the limited capability of the diffusion model to capture and represent the continuous properties of these images. While \emph{MultiDiffusion} demonstrates effectiveness in generating ordinary panoramic images, it also encounters difficulties in ensuring continuity between the leftmost and rightmost sides of the generated images. However, by incorporating our designed \emph{StitchDiffusion} method, the customized model accurately synthesizes seamless 360-degree panoramic images.

% \begin{figure*}[t]
% \centering
% \vspace{-2mm}
% \includegraphics[height=3.5cm]{sliding_distance.pdf}
% \caption{Ablation study on the horizontal sliding distance. The corresponding text prompt is `$V^*$, castle, a beautiful artwork illustration'. When the horizontal sliding distance $\omega$ in the \emph{StitchDiffusion} is 128, the content in the \textcolor{green}{green solid box} is more consistent and seamless than those in the \textcolor{blue}{blue} and \textcolor{red}{red} solid boxes for the sliding distances of 256 and 512.}
% \label{fig:sliding_distance}
% \vspace{-3mm}
% \end{figure*}

% \begin{figure}
% \centering
% \vspace{-3mm}
% \includegraphics[height=6.4cm]{limitation3.pdf}
% \caption{Illustration of limitations. Note that the two images are not conducted the final global cropping yet. If there is dissimilar content between the two sub-regions denoted by the \textcolor{green}{green solid line} (e.g., plant and wall) in the generated image without additional denoising, the corresponding region in the synthesized image with additional denoising exhibits inconsistent content.} 
% \vspace{-3mm}
% \label{fig:limitation}
% \end{figure}

\noindent
\textbf{Trigger Word Ablation.} 
To investigate the impact of the trigger word $V^*$ on the synthesis process, we conducted a comparison between images generated by our method with and without the trigger word included in the input text prompts. The visual results are shown in Figure \ref{fig:trigger}. We can see that the trigger word $V^*$ plays a crucial role in the generated image. When the trigger word is omitted from the text prompt, the resulting image fails to encompass the entire field of view spanning 360 degrees horizontally and 180 degrees vertically. Conversely, when the text prompt includes the trigger word, our customized model with \emph{StitchDiffusion} successfully produces 360-degree panoramas.

\noindent
\textbf{Number of Pre-Denoising Times Ablation.} In our \emph{StitchDiffusion} method, we perform pre-denoising operations twice on the stitch block at each time step $t$ of the denoising process, as depicted in Figure \ref{fig:stitch}. To explore the impact of the number of pre-denoising operations on the synthesized images, we conducted an experiment comparing the results obtained with different numbers of pre-denoising operations at each time step $t$. The results are presented in Figure \ref{fig:denoising_time}. We can see that when we only conduct one pre-denoising operation, despite the leftmost and rightmost sides of the generated image are continuous, there is a local content inconsistency exampled by the \textcolor{red}{red solid box} in the image. However, by conducting two or three pre-denoising operations on the stitch block, our customized model effectively generates high-quality 360-degree panoramic images without noticeable local content inconsistency. For a higher computational efficiency, we adopt the approach of performing pre-denoising operations twice in our method.

\section{Conclusion}

In this study, we have explored the customization of a T2I diffusion model for generating 360-degree panoramas. Our approach involved the establishment of a paired image-text dataset called \emph{360PanoI}, followed by fine-tuning latent diffusion using LoRA. However, the fine-tuned diffusion model alone falls short in ensuring the continuity between the leftmost and rightmost sides of the generated images. To address this limitation, we proposed a method called \emph{StitchDiffusion}, which successfully enables the customized diffusion model to synthesize seamless 360-degree panoramas. Through extensive experiments, we have verified the effectiveness of the proposed method and demonstrated that our customized diffusion model exhibits exceptional generalization ability, producing diverse and high-quality 360-degree panoramic images even in previously unseen scenes. The applications of our work are vast, particularly in fields such as indoor design, game and VR content creation, where the utilization of 360-degree panoramas is prevalent. Moreover, the \emph{360PanoI} dataset we collected will be beneficial for any future investigations into 360-degree panoramic images.

%%%%%%%%% REFERENCES
{\small
\bibliographystyle{ieee_fullname}
\bibliography{stitchdiffusion}
}

\appendix

%%%%%%%%% BODY TEXT
\section{Supplementary Content}

This supplementary material begins by presenting additional ablation studies. Next, we showcase sample images from various scenes in our collected \emph{360PanoI} dataset and generation process of their text prompts. Finally, more images synthesized using different methods are shown.

\subsection{Additional Ablation Studies}

\begin{figure}[t]
\centering
\includegraphics[height=8.8cm]{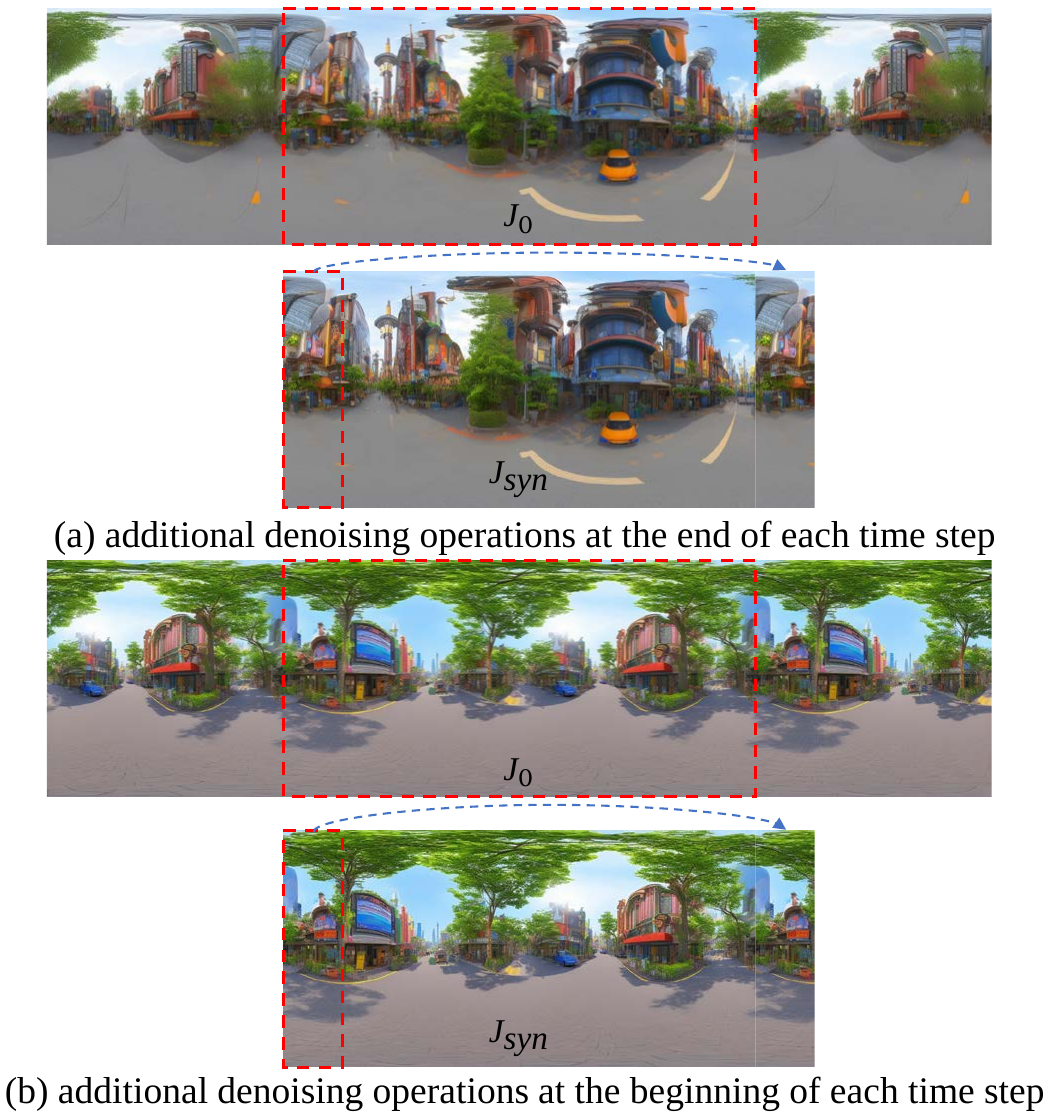}
\caption{Ablation study on the order of additional denoising operations. The corresponding text prompt is `$V^*$, fantastical street, cinematic, anime style'. $J_0$ and $J_{syn}$ denote the clear denoised image and the final result, respectively. The leftmost and rightmost sides of $J_{syn}$ are not continuous when the additional denoising operations are performed twice at the end of denoising time step $t$. }
\label{fig:order}
\end{figure}

\noindent
\textbf{Order of Additional Denoising Operations Ablation.} 
In our proposed \emph{StitchDiffusion} method, we additionally perform pre-denoising operations twice on the stitch block at each time step $t$ during the denoising process. In this experiment, we investigate the effect of the order in which these additional denoising operations are conducted, specifically at the beginning and at the end of denoising time step $t$. The corresponding results are shown in Figure \ref{fig:order}. We can observe that: (1) if the additional denoising operations are performed twice on the stitch block at the end of denoising time step $t$, the synthesized image does not form a seamless 360-degree panorama; (2) however, when the additional denoising operations are conducted twice on the stitch block at the beginning of denoising time step $t$, the customized diffusion model effectively generates a seamless 360-degree panoramic image.

\begin{figure*}[t]
\centering
\includegraphics[height=3.5cm]{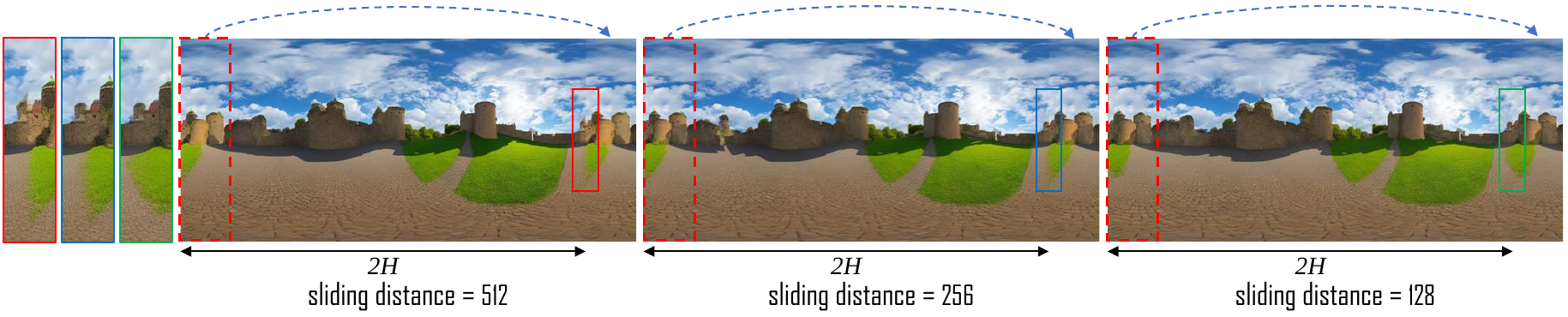}
\caption{Ablation study on the horizontal sliding distance. The corresponding text prompt is `$V^*$, castle, a beautiful artwork illustration'. When the horizontal sliding distance $\omega$ in the \emph{StitchDiffusion} is 128, the content in the \textcolor{green}{green solid box} is more consistent and seamless than those in the \textcolor{blue}{blue} and \textcolor{red}{red} solid boxes for the sliding distances of 256 and 512.}
\label{fig:sliding_distance}
\end{figure*}

\noindent
\textbf{Horizontal Sliding Distance Ablation.} To study the effect of horizontal sliding distance $\omega$ between adjacent cropped patches on the synthesized images, a comparison was carried out using various sliding distances. The visual results are presented in Figure \ref{fig:sliding_distance}. There is a noticeable seam in the middle of the region indicated by the \textcolor{red}{red solid box} when the sliding distance $\omega$ is set to 512. By reducing the sliding distance to 256, the noticeable seam is improved, but the local content inconsistency (ground and grass) still exists in the region now represented by the \textcolor{blue}{blue solid box}. Finally, with a sliding distance of $\omega$ set to 128, the content within the region now marked by the \textcolor{green}{green solid box} exhibits seamless and consistent integration.

\begin{figure*}[htbp]
\centering
\includegraphics[height=12.1cm]{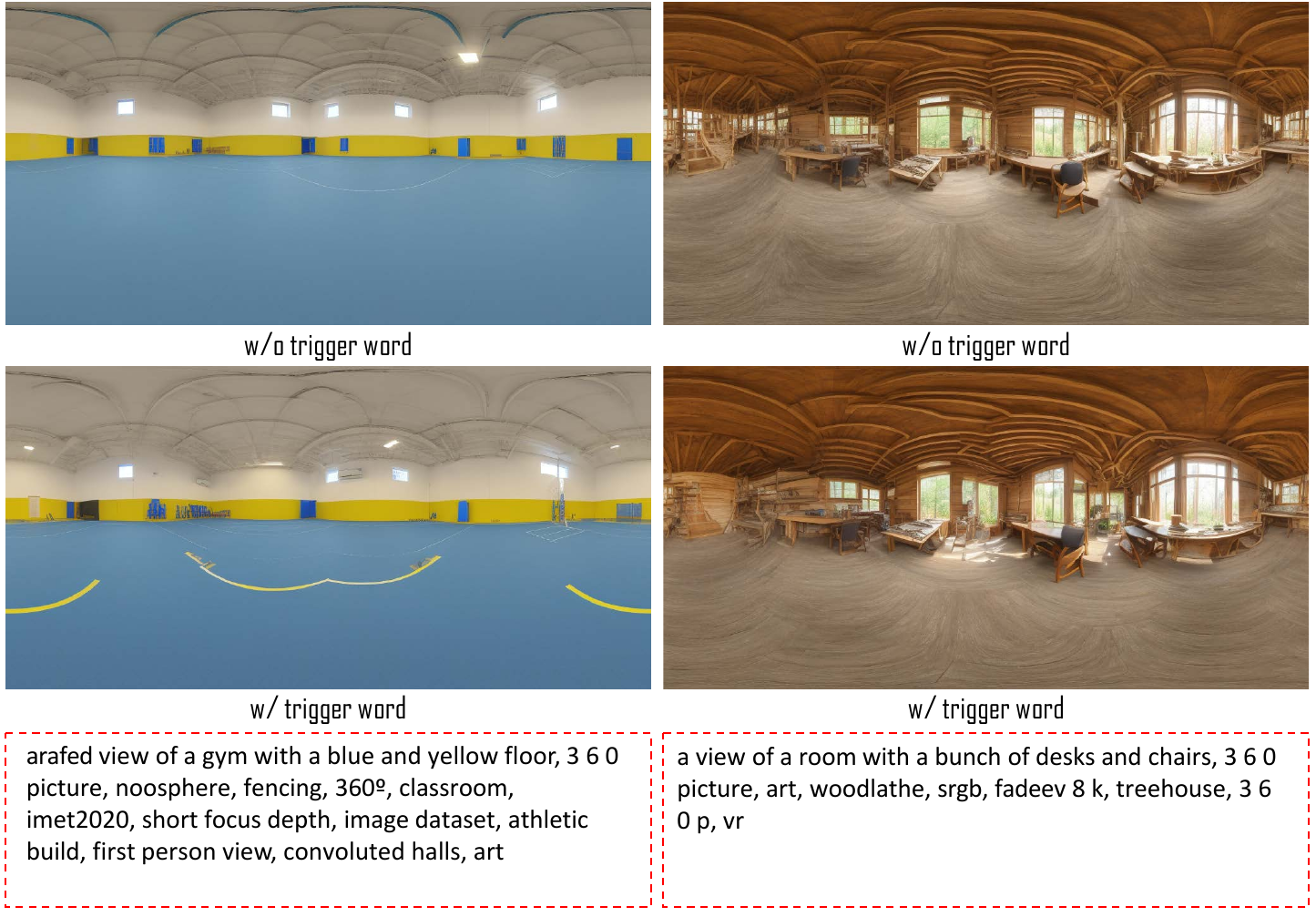}
\caption{Ablation study on the effectiveness of the trigger word $V^*$ if poor tags in text prompts are not filtered out. We collected two real 360-degree panoramas from Poly Haven \cite{polyhaven}, which are independent of the \emph{360PanoI} dataset, and utilized BLIP \cite{blip} to create corresponding text prompts denoted by the \textcolor{red}{red dashed box}. It is evident that without filtering out poor tags like `3 6 0 picture', the trigger word cannot effectively control our method's generation of 360-degree panoramas.}
\label{fig:trigger2}
\end{figure*}

\noindent
\textbf{Poor Tags Ablation.} To assess the impact of poor tags within the input text prompt on the trigger word's effectiveness in controlling the image generation, we employed BLIP \cite{blip} to get the corresponding text prompts from real 360-degree panoramas. Subsequently, we conducted a comparison between images generated using these poor-tags-included text prompts with and without our trigger word. As shown in Figure \ref{fig:trigger2}, the presence of these poor tags in the text prompt hinders our trigger word's ability to control the generation of 360-degree panoramas by our method.

\begin{figure*}[htbp]
\centering
\includegraphics[width=\textwidth]{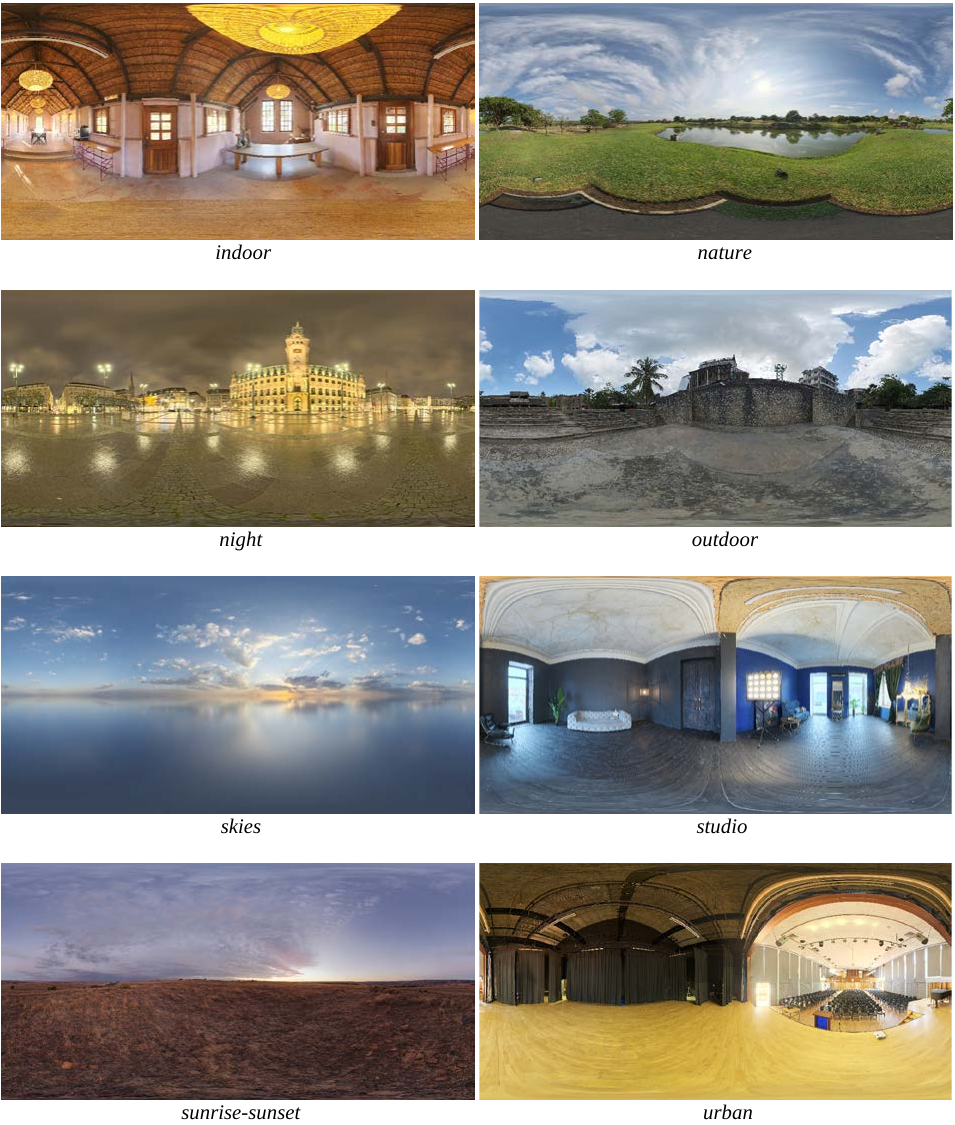}
\caption{Sample images depicting the eight scenes contained within our collected \emph{360PanoI} dataset are presented. In order to fine-tune a text-to-image diffusion model for customizing 360-degree panoramas, we utilize the entire set of 120 panoramic images from the dataset along with their corresponding text prompts acquired from BLIP \cite{blip}.}
\label{fig:sample}
\end{figure*}

\begin{figure*}[htbp]
\centering
\includegraphics[width=\textwidth]{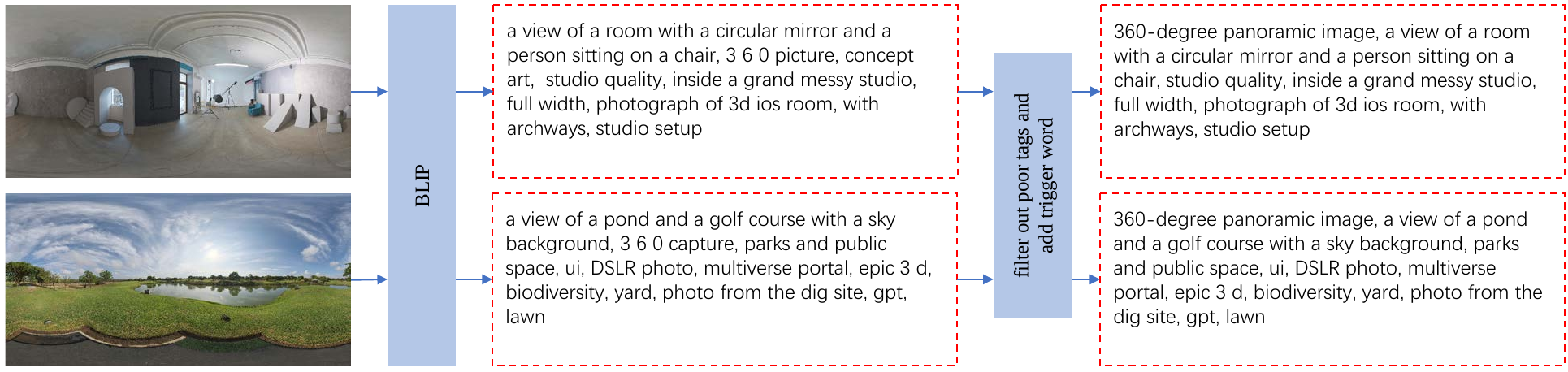}
\caption{Diagram to show the generation process of text prompts in our \emph{360PanoI} dataset using BLIP \cite{blip}. For the collected 120 360-degree panoramas, we employ BLIP to produce their text prompts. Then, we filter out poor tags such as `3 6 0 picture' and introduce a trigger word `360-degree panoramic image' into each text prompt, resulting in the final text prompts in our \emph{360PanoI} dataset. Note that we only show 2 images from the 120 collected panoramas for illustration.}
\label{fig:text}
\end{figure*}

\begin{figure*}[htbp]
\centering
\includegraphics[height=10cm]{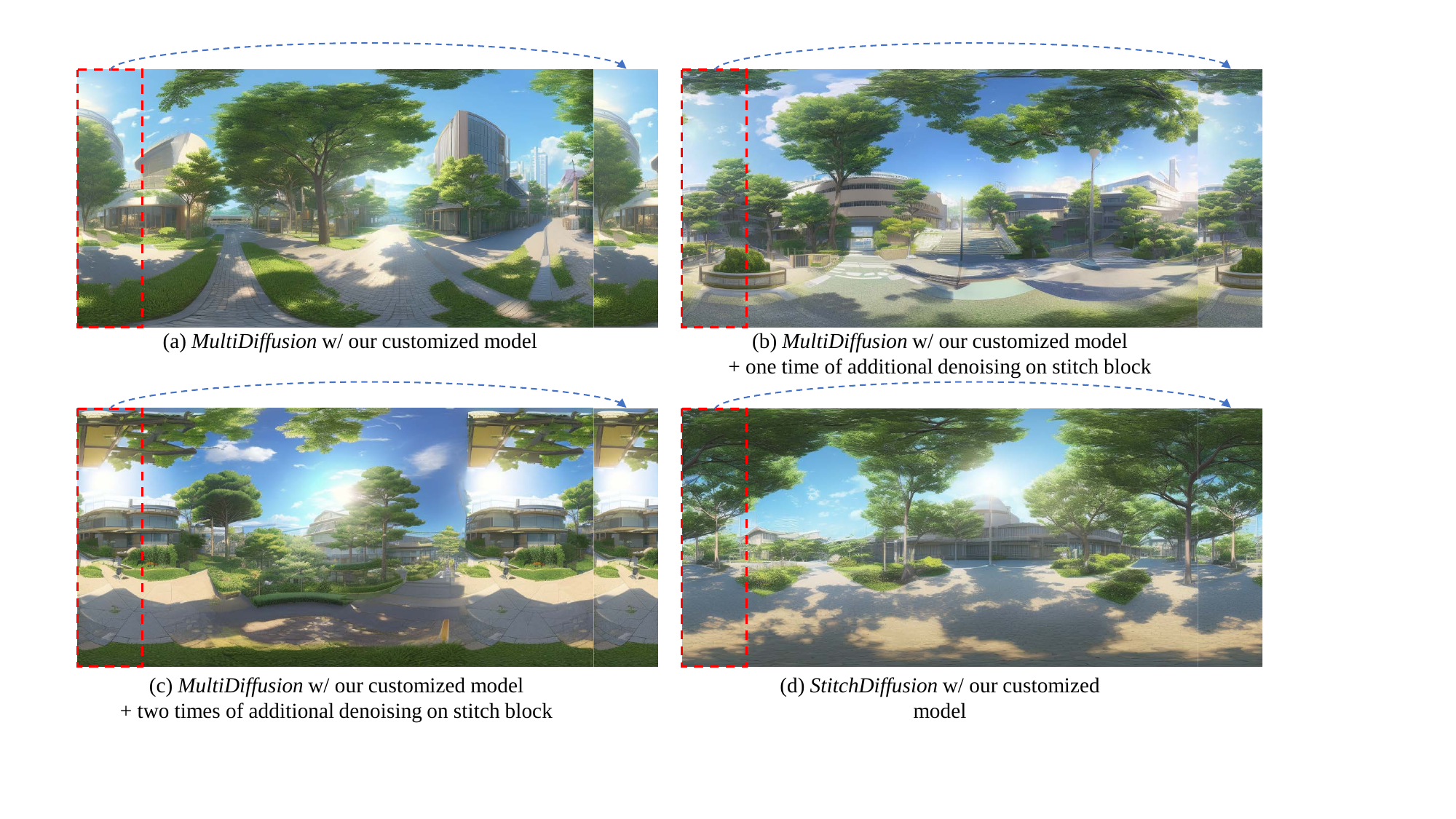}
\caption{Visual results of different schemes. The corresponding text prompt for these generated images is `360-degree panoramic image, campus, unreal engine, studio quality, japanese anime style, anime'. Since the size of the cropped patch in \emph{MultiDiffusion} \cite{bar2023multidiffusion} is $512\times512$, the stitch block here consists of the leftmost ($512\times256$) and rightmost ($512\times256$) regions in the image. We can see that the combination of \emph{MultiDiffusion} and our customized model in (a) cannot generate a seamless 360-degree panorama. Even with one time of additional denoising applied to the stitch block in (b), or two times of additional denoising applied to the stitch block in (c), the results remains unseamless. In contrast, our method in (d) successfully synthesizes a seamless and plausible 360-degree panorama that aligns with the text prompt.}
\label{fig:multi_stitch}
\end{figure*}

\subsection{Sample Images and Their Text Prompts}

The 360-degree panoramas in our collected \emph{360PanoI} dataset have been sourced from Poly Haven \cite{polyhaven}. To display the different scenes within our dataset, we randomly select one image from each scene. The corresponding sample images, with a resolution of $512\times1024$, are illustrated in Figure \ref{fig:sample}. Despite the \emph{360PanoI} dataset only contains 8 scenes from the real world, it is important to note that the visual results presented in the main manuscript demonstrate the generalization capability of our customized diffusion model, utilizing the proposed \emph{StitchDiffusion} technique, to generate 360-degree panoramas encompassing a wide variety of scenes unseen in the dataset. In addition, we provide a diagram in Figure \ref{fig:text} to demonstrate the generation process of the corresponding text prompts for these 360-degree panoramas within the \emph{360PanoI} dataset.

\subsection{More Visual Results}

To highlight the distinctions between \emph{MultiDiffusion} \cite{bar2023multidiffusion} and our proposed \emph{StitchDiffusion}, we present a visual comparison between various schemes of \emph{MultiDiffusion} and our  \emph{StitchDiffusion} in Figure \ref{fig:multi_stitch}. We can see that the combination of \emph{MultiDiffusion} and our customized model in (a) fails to generate a seamless 360-degree panorama. Even with one time of additional denoising applied to the stitch block in (b), or two times of additional denoising applied to the stitch block in
(c), the results remains unseamless. In contrast, our method in (d) successfully synthesizes a seamless and plausible 360-degree panorama corresponding to the text prompt.

\begin{figure*}[htbp]
\centering
\includegraphics[height=20.6cm]{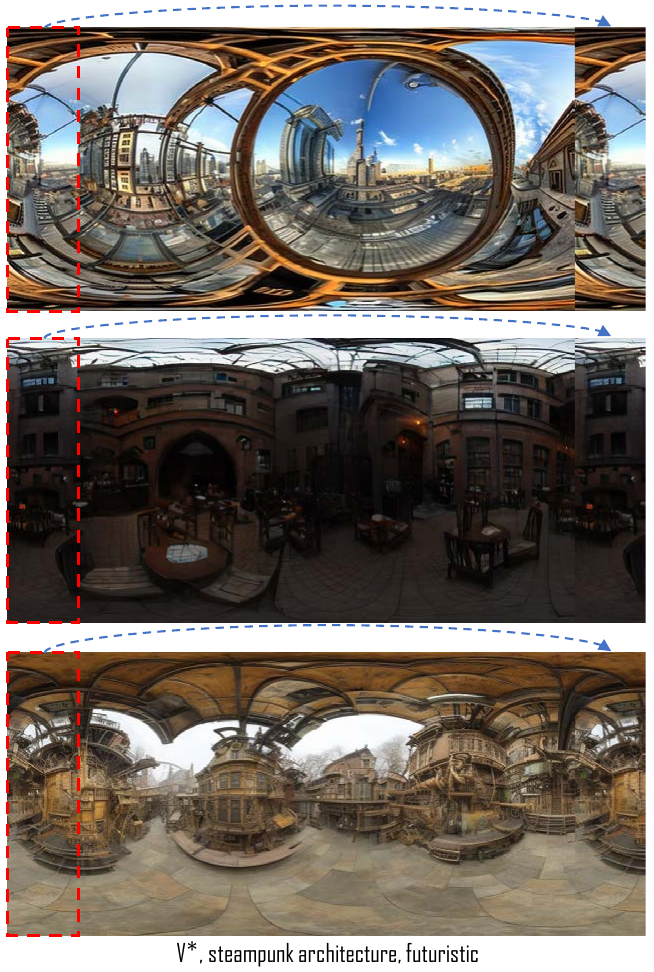}
\caption{Visual comparison among \emph{MultiDiffusion} \cite{bar2023multidiffusion} combined with latent diffusion  (the first row), Text2Light \cite{text2light} (the second row) and our method (the third row).  To demonstrate the discontinuity or continuity between the leftmost and rightmost sides of the generated image, we copy the leftmost area ($512\times128$) represented by the \textcolor{red}{red dashed box} and paste it onto the rightmost side of the image. Our method generates a photorealistic and seamless 360-degree panoramic image compared to the other two methods.}
\label{fig:comparison1}
\end{figure*}

\begin{figure*}[htbp]
\centering
\includegraphics[height=20.2cm]{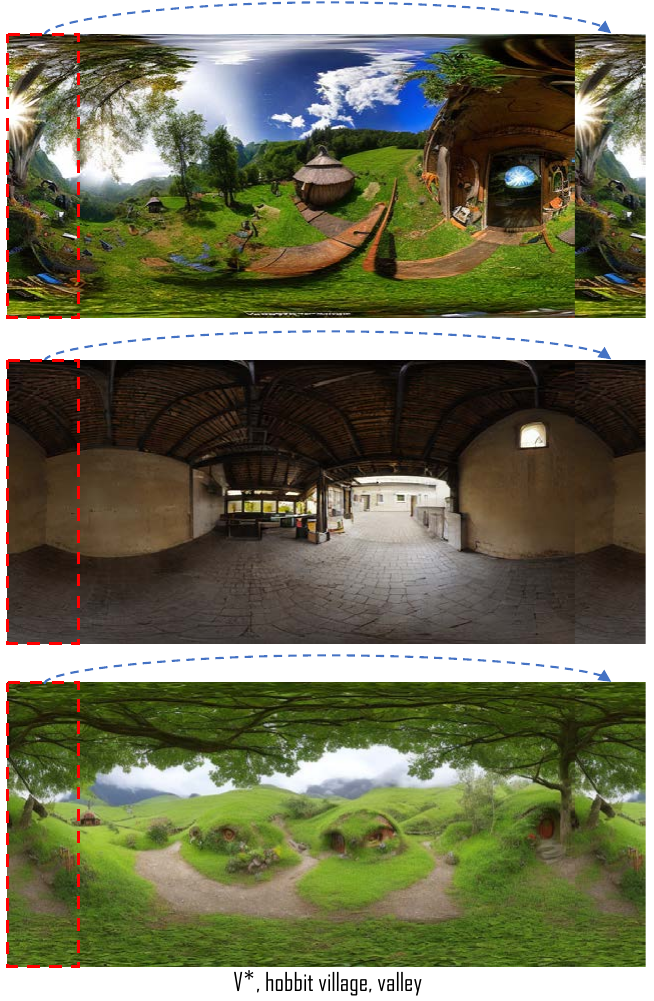}
\caption{Visual comparison among \emph{MultiDiffusion} \cite{bar2023multidiffusion} combined with latent diffusion  (the first row), Text2Light \cite{text2light} (the second row) and our method (the third row). To display the discontinuity or continuity between the leftmost and rightmost sides of the generated image, we copy the leftmost area ($512\times128$) represented by the \textcolor{red}{red dashed box} and paste it onto the rightmost side of the image. Compared to the two other approaches, our method excels in generating visual appealing and plausible 360-degree panoramas aligned with the input text prompts.}
\label{fig:comparison2}
\end{figure*}

\begin{figure*}
\centering
\includegraphics[width=\textwidth]{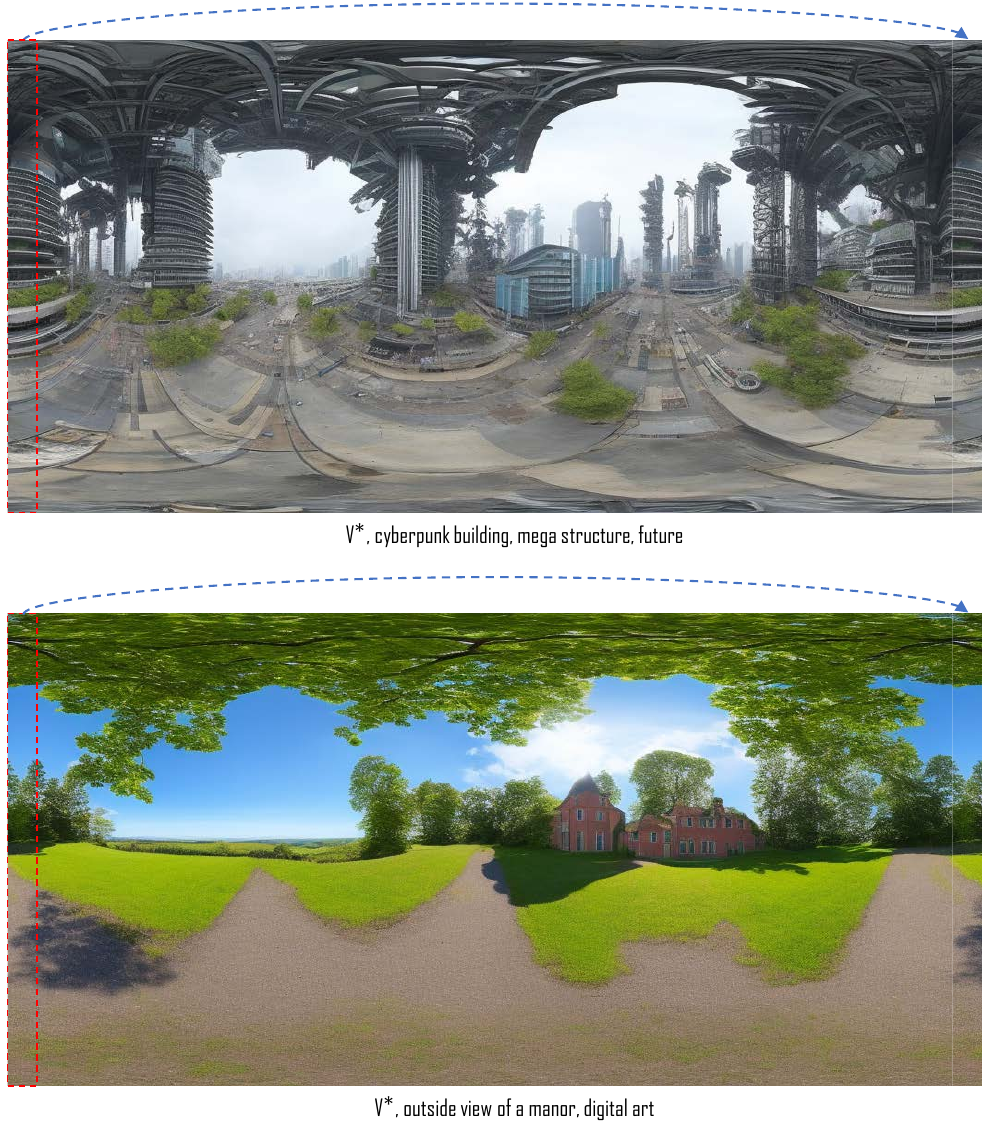}
\caption{The images generated by our method showcasing the themes of `cyberpunk' and `manor' are presented.  To show the discontinuity or continuity between the leftmost and rightmost sides of the generated image, we copy the leftmost area ($512\times32$) represented by the \textcolor{red}{red dashed box} and paste it onto the rightmost side of the image. By carefully observing the illustrations, we can see that our method successfully captures the essence of the `cyberpunk' and `manor' themes, generating visually appealing 360-degree panoramic images.}
\label{fig:diverse3}
\end{figure*}

\begin{figure*}
\centering
\includegraphics[width=\textwidth]{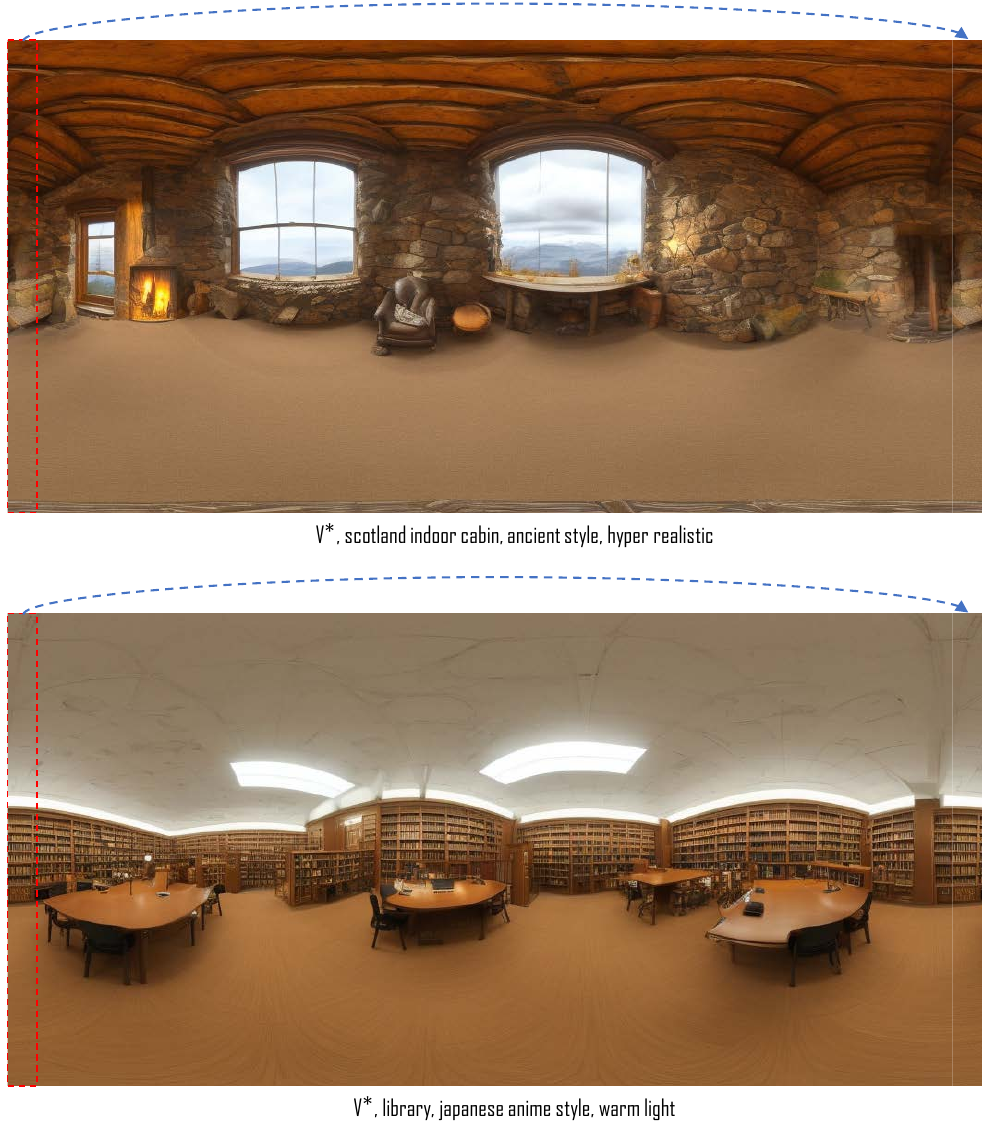}
\caption{Illustration of `cabin' and `library' generated by our method.  To display the discontinuity or continuity between the leftmost and rightmost sides of the generated image, we copy the leftmost area ($512\times32$) represented by the \textcolor{red}{red dashed box} and paste it onto the rightmost side of the image. The continuity between the leftmost and rightmost sides of the synthesized images is effectively maintained, ensuring a seamless transition and enhancing the overall immersive experience for viewers.}
\label{fig:diverse4}
\end{figure*}

To further demonstrate the superiority of our method in generating 360-degree panoramas, we provide additional comparison results involving our method, latent diffusion  combined with \emph{MultiDiffusion} \cite{bar2023multidiffusion}, and Text2Light \cite{text2light}, shown in Figure \ref{fig:comparison1} and Figure \ref{fig:comparison2}. Our method outperforms the other two methods by producing seamless and plausible 360-degree panoramic images that correspond to the input text prompts. Moreover, to showcase the excellent generalizability of our proposed method, we present more synthesized 360-degree panoramas depicting various scenes in Figure \ref{fig:diverse3} and Figure \ref{fig:diverse4}.

\end{document}